\documentclass[runningheads]{llncs}
\usepackage{graphicx,calc} 
\usepackage{algorithm} 
\usepackage[algo2e,ruled,linesnumbered]{algorithm2e} 
\usepackage[]{algpseudocode}
\usepackage{bbm}

\usepackage{hyperref}
\usepackage{amsmath}
\usepackage{amssymb} 
\usepackage{hhline}
\usepackage{multirow}
\usepackage{booktabs} 
\usepackage{threeparttable}
\usepackage{multirow}
\usepackage[T1]{fontenc}
\usepackage{caption}
\usepackage{subcaption}
\usepackage{tikz}
\usepackage{orcidlink}
\usepackage[misc]{ifsym}

\usepackage{xcolor}

\DeclareMathOperator*{\argmax}{argmax} 

\newcommand{\plm}{\textsc{Plm}}

\newcommand{\plmkm}{\plm{}-\textsc{km}}
\newcommand{\kmeans}{K-Means}
\newcommand{\entropy}{\textsc{Entropy}}
\newcommand{\badge}{\textsc{Badge}}
\newcommand{\bald}{\textsc{Bald}}
\newcommand{\alps}{\textsc{Alps}}
\newcommand{\call}{\textsc{Cal}}
\newcommand{\actune}{\textsc{AcTune}}
\newcommand{\random}{\textsc{Random}}
\newcommand{\real}{\textsc{Real}}
\newcommand{\al}{\textsc{al}}

\newcommand{\roberta}{\textsc{RoBERTa}}

\newcommand{\Du}{\mathcal{D}_u}
\newcommand{\Dl}{\mathcal{D}_l}

\newcommand{\cls}{\texttt{[CLS]}}
\newcommand{\lift}{\texttt{lift}}

\newcommand{\toyScale}{0.8}
\newcommand{\toyPseRect}{\includegraphics[scale=\toyScale]{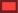}}
\newcommand{\toyPseTri}{\includegraphics[scale=\toyScale]{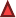}}%
\newcommand{\toyPseTriBlue}{\includegraphics[scale=\toyScale]{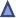}}%
\newcommand{\toyPseCirBlue}{\includegraphics[scale=\toyScale]{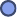}}%

\newcommand{\sst}{\textsc{sst-$2$}}
\newcommand{\ag}{\textsc{agnews}}
\newcommand{\pub}{\textsc{pubmed}}
\newcommand{\snips}{\textsc{snips}}
\newcommand{\stov}{\textsc{stov}}

\newcommand\blfootnote[1]{%
  \begin{NoHyper}%
  \begingroup
  \renewcommand\thefootnote{}\footnote{#1}%
  \addtocounter{footnote}{-1}%
  \endgroup
  \end{NoHyper}%
}

\begin{document}
\title{ \real{}: A Representative Error-Driven Approach for Active Learning}
%
\institute{}

\author{Cheng Chen$^1$\orcidlink{0009-0006-6805-5894} 
Yong Wang$^{2}$(\Letter)\orcidlink{0000-0002-0092-0793}
Lizi Liao$^2$\orcidlink{0000-0002-9973-3305}
Yueguo Chen$^{1}$\orcidlink{0000-0002-2239-4472}
Xiaoyong Du$^1$\orcidlink{0000-0002-5757-9135}\\
    $^1$Renmin University of China \enspace
    $^2$Singapore Management University\\
  \texttt{\{chchen, chenyueguo, duyong\}@ruc.edu.cn} \\
  \texttt{\{yongwang, lzliao\}@smu.edu.sg}
  }
\authorrunning{Cheng Chen et al.}

\toctitle{\real{}: A Representative Error-Driven Approach for Active Learning}
\tocauthor{Cheng~Chen; Yong~Wang; Lizi~Liao; Yueguo~Chen; Xiaoyong~Du}

%
\maketitle              
\begin{abstract}

Given a limited labeling budget, active learning (\al{}) aims to sample the most informative instances from an unlabeled pool to acquire labels for subsequent model training.
To achieve this, \al{} typically measures the informativeness of unlabeled instances based on uncertainty and diversity.
However, it does not consider erroneous instances with their neighborhood error density,
which have great potential to improve the model performance. 
To address this limitation, we propose \real{}, 
a novel approach to select data instances with
\underline{R}epresentative \underline{E}rrors for \underline{A}ctive \underline{L}earning. It identifies minority predictions as \emph{pseudo errors} within a cluster and allocates an adaptive sampling budget for the cluster based on estimated error density.
Extensive experiments on five text classification datasets demonstrate that \real{} consistently outperforms all best-performing baselines regarding accuracy  and F1-macro scores across a wide range of hyperparameter settings.
Our analysis also shows that \real{} selects the most representative pseudo errors that match the distribution of ground-truth errors along the decision boundary.
Our code is publicly available at \url{https://github.com/withchencheng/ECML_PKDD_23_Real}.
\blfootnote{(\Letter)The corresponding author.}
\end{abstract}
\section{Introduction}
\begin{figure}
    \centering
    \includegraphics[scale=\toyScale]{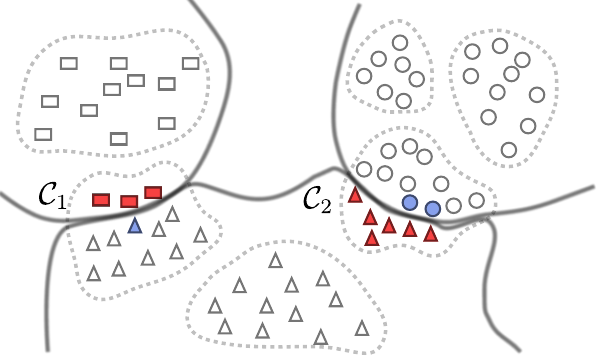} 
    \caption{\small An illustrative example of \real{}.
      The solid thick lines denote the model decision boundaries, separating data points into three predicted  classes (rectangles, circles, and triangles).
      The dashed irregular circles are clusters of data points.
      \real{} samples the minority predictions as \emph{pseudo errors} (\toyPseRect{}  \toyPseTri{}) in cluster $\mathcal{C}_1$ and $\mathcal{C}_2$ for labeling.
      If our budget exceeds the number of pseudo errors, \real{} will pick blue instances (\toyPseTriBlue{} \toyPseCirBlue{}) where the model has the least confidence in its predictions.
    }%
    \label{fig:real_toy}
    \vspace{-0.3cm}
\end{figure}
Labeling data for machine learning is costly, and the budget on the amount of labels we can gather is often limited.
Therefore, it is crucial to make the training process of machine learning models more label-efficient,
especially for applications where labels are expensive to acquire.
Active learning (\al{}) is to select a small amount of the most informative instances from an unlabeled pool, aiming to maximize the model performance gain when using the selected instances (labeled) for further training.
Identification of the most informative instances from the unlabeled data pool is critical to the success of \al{}.

The \al{} techniques can be classified into three groups: uncertainty-based, diversity-based, and hybrid methods.
Uncertainty-based methods select instances whose prediction probability is more evenly distributed over classes~\cite{lewis1994heterogeneous,lewis1995sequential}, instances with a larger expected loss/gradient~\cite{yoo2019learning,ash2020deep}, or those closer to decision boundaries~\cite{ru2020active,ducoffe2018adversarial}.
However, solely relying on instance-level uncertainty metrics may cause redundancy in samples~\cite{yu2022actune}.
Hence, diversity-based methods try to mitigate the redundancy problem by selecting a small but diverse set of data instances to represent \emph{the whole unlabeled pool}~\cite{kim2022core-set}.
However, they ignore the fact that training on errors is more label-efficient~\cite{choi2021vab}.
Hybrid methods try to select instances that are both uncertain and diverse\cite{yuan2020alps,yu2022actune}.
Our proposed method belongs to the hybrid category.
Our novelty is to seek representativeness for the \emph{errors} rather than the \emph{whole unlabeled pool}, by selecting instances with a larger error probability and higher neighborhood error density.
Fig.~\ref{fig:real_toy} shows an illustrative example of our method. Specially, we first cluster the unlabeled instances by their representations.
The majority prediction in a cluster is expected to be correct even with limited labeled training data~\cite{aharoni2020unsupervised,sia2020tired,chen2020simple}, owing to the strong representation power of the pretrained models for images~\cite{he2020momentum} or texts~\cite{liu2019roberta}.
Also, it is common for \al{} to achieve a decent test accuracy after the warm-up training on the initial limited labeled data~\cite{margatina2021active,yu2022actune}.
Consequently, we treat the majority prediction in a cluster as the \emph{pseudo label} for all the instances in the cluster.
We call instances in a cluster whose predictions disagree with the cluster \emph{pseudo label}  as \emph{pseudo errors}. More pseudo errors with a lower prediction probability (larger disagreement) on the pseudo label will bring a larger sampling budget to its affiliated cluster. In this way, we emphasize dense areas of errors, and thus adaptively select more representative errors.

To our best knowledge, \real{} is the first approach to sample representative errors to achieve label-efficient active learning.
By taking text classification as an example application, we demonstrate the effectiveness of \real{}.
In summary, the major contributions of this paper can be summarized as follows:
\begin{itemize}
    \item We propose a new \al{} sampling algorithm, \real{}, that explores selecting representative  errors from the unlabeled pool.
    \item We show \real{} consistently beats all the best-performing baselines on five text classification benchmark datasets
    in terms of both accuracy and F1-macro scores.
    \item We empirically investigate error distribution and find that 1) most errors are distributed along the decision boundary; 2) the distribution of selections made by \real{} align well with that of ground-truth errors.
\end{itemize}

\section{Preliminaries}
\subsection{Related Work}
\subsubsection*{Uncertainty-based}
Uncertainty-based sampling is to sample the most uncertain instances for model training.
Three  classical metrics for the uncertainty of model prediction probabilities are: entropy~\cite{lewis1994heterogeneous,lewis1995sequential}, least confidence~\cite{lewis1995sequential,li2006confidence}, and smallest margin~\cite{roth2006margin}.
Recent research studies take the expected loss~\cite{yoo2019learning}, expected generalization error reduction~\cite{konyushkova2017learning},
or distance to the decision boundary~\cite{ru2020active} as surrogates for uncertainty.
\call{}~\cite{margatina2021active} selects contrastive examples that are similar in the feature space of pre-trained language model (\plm{}) and
 maximally different in the output probabilities.
Unlike \call{} which ignores  the correctness of sampled instances, our method aims to mine the yet-to-be errors.
OPAL \cite{krempl2015OPAL} computes the expected misclassification loss reduction, but is limited to binary classification using the outdated Parzen window classifier.

\vspace{-1.5em}
\subsubsection*{Diversity-based} Diversity-based sampling aims to maximize the diversity of sampled instances.
Cluster-Margin~\cite{citovsky2021batch} selects a diverse of instances with the smallest margin  using  hierarchical agglomerate clustering.
Sener and Savarese~\cite{sener2018active} proposed a coreset approach to find a representative subset from the unlabeled pool.
Kim~\textit{et al.}~\cite{kim2022core-set} assessed the density of unlabeled pool and selected diverse samples mainly from regions of low density.
Meanwhile, generative adversarial learning \cite{gissin2019discriminative} is applied in \al{} as a binary classification task. They trained an adversarial classifier to confuse data from the training set and that from the pool.
However, our method cares about the density of errors rather than the whole unlabeled pool.

\vspace{-1.5em}
\subsubsection*{Hybrid}
Hybrid \al{} methods try to combine uncertainty and diversity sampling.
Such a combination can be achieved by 
meta learning~\cite{baram2004online,hsu2015active} and reinforcement learning~\cite{fang2017learning,liu2018-learning-actively},
which  automatically learn a sampling strategy in each \al{} round instead of using a fixed heuristic.
\badge{}~\cite{ash2020deep} and \alps{}~\cite{yuan2020alps} both compute uncertainty representations of instances and then cluster them.
\badge{} transforms data into gradient embeddings that encode the model confidence and then apply \kmeans{}++~\cite{arthur2007km}.
\alps{} first utilizes the  self-supervision loss of \plm{} as uncertainty representation.
\actune{}~\cite{yu2022actune} uses weighted \kmeans{}  clustering to find highly uncertain regions. 
The \kmeans{}  clustering in \actune{} is weighted by some uncertainty measure, e.g., entropy or \call{}~\cite{margatina2021active}.
\actune{} deliberately tries to separate the uncertain regions from the confident regions via weighted clustering, with an implicit assumption that the two kinds of regions are separable.

\subsection{Problem Definition}
We take text classification as an example to illustrate the core idea of our approach.
Given a small labeled set $ \Dl = \{(\mathrm{x}_i, y_i)\}_{i=1}^L$  (warm-up dataset) for initial model training and a large unlabeled data pool $ \Du = \{(\mathrm{x}_i,)\}_{i=1}^U$,
where $\mathrm{x}_i$ is the $i-$th input instance (e.g., the token sequence for text classification), 
$y_i \in \{1, \ldots ,Y\} $ is the target label, and $\Dl \ll \Du$,
we want to select and obtain the labels of the most informative instances in $ \Du $ for training model $\mathcal{M}$, 
so that the performance of $\mathcal{M}$ can be maximized given a fixed labeling budget $B$
\footnote{Following the convention in machine learning community~\cite{lewis1994heterogeneous,konyushkova2017learning,liu2018-learning-actively,citovsky2021batch}, we ignore the cognitive difference for labeling different instances studied in the HCI community\cite{chung2021understanding,muller2021designing},
and assume the labeling cost is $1$ for every instance. For example, if our total labeling budget is  $B=800$ and we have $T=8$ rounds of \al{}, then $b=100$ is the budget per round.}. %
$\mathcal{M}$ is trained iteratively.
Suppose there are $T$ \al{} rounds in total, then the budget for each round is $b=B/T$.
In each \al{} round, a sampling function $\alpha(\Du, \mathcal{M})$ selects $b$ samples from
$\Du$ based on the previously learned model $ \mathcal{M}$,
and then moves the labeled $b$ samples into $\Dl$.
Model $\mathcal{M}$ is trained on the updated  $\Dl$ and then evaluated on a hold-out test set.
The \al{} process terminates when the total budget $B$ is exhausted or the model performance is good enough.
The core of an \al{} method
is to study the sampling function $\alpha$.

\section{Representative Pseudo Errors}
\subsection{Overview of Representative Error-Driven Active Learning}
We aim to explore one critical research question for active learning:
how will \emph{learning from errors} improve the active learning accuracy for models?
Intuitively, sampling more errors for model training will prevent the model from making the same mistakes on the test set, thus improving the test accuracy. 
Errors bring larger loss values, making them more informative for model training~\cite{yoo2019learning}. 
Though one existing work \cite{choi2021vab} tries to directly calculate the erroneous probability for some image using the Bayesian theorem, it ignores the density of errors.
Other than computing a single unlabeled instance's erroneous probability, we develop a sense of representativeness for the selected errors into our approach.

The \al{} process starts from training the model $\mathcal{M}$ on the initial labeled data set
$ \Dl^{(0)} $.
Formally, we minimize the average cross entropy loss $\ell$ for all the instances in  $\Dl^{(0)}$:
\begin{equation}
    \label{eq:train_dl}
    \min_\theta \frac{1}{|\Dl^{(0)}|}  \sum_{(\mathrm{x}_i, y_i) \in \Dl}  \ell ( \mathcal{M}(\mathrm{x}_i, \theta^{(0)}), y_i) .
\end{equation}

In each of the following \al{} rounds, \real{} (Algorithm~\ref{algo:real}) selects a set of representative errors $Q$ consisting of $b$ instances from $\Du$, obtains their labels, and then adds $Q$ into $\Dl$ for subsequent model training.
Algorithm~\ref{algo:real} consists of two components: \emph{pseudo error identification} and \emph{adaptive sampling of representative errors}.

\subsection{Pseudo Error Identification}

The first challenge is to select instances from $\Du$ where the model $\mathcal{M}$ makes mistakes, which is non-trivial since we do not have access to the ground truth labels before selection.
However, prior studies~\cite{aharoni2020unsupervised,sia2020tired,zhang2022neural} have shown that
\plm{} can effectively learn the sentence representations and support accurate classifications very well by simply clustering the embedded representations of sentences~\cite{sia2020tired}.
Also, it is commonly-seen that active learning will be employed after the machine learning models have achieved a reasonable performance~\cite{margatina2021active,yu2022actune}. 
Building upon these facts,
it is safe to expect the majority prediction in a cluster by the \plm{} has a high probability to be the ground truth label, even with a small amount of training data.
Thus,
we assume that the majority prediction is the correct label for all the instances in the cluster.
Our preliminary experiments also show a relatively high and stable accuracy of our pseudo-label assignment strategy, i.e., over 0.80 for all the chosen datasets.
Since the majority prediction is treated as the pseudo label to each cluster,
pseudo errors are defined as those instances whose predictions disagree with the majority prediction in each cluster.
As will be shown in Section \ref{sec:experiment_results}, the sampled pseudo errors usually have higher error rates when compared with ground truths, indicating that such a way of defining pseudo labels and pseudo errors is effective.

In round $t ~(1 \le t \le T)$ of active learning, we first obtain the representations of instances in $\Du$
by feeding them into model $\mathcal{M}$'s encoder $\Phi(.)$.
Specifically, we only take the \cls{} token embedding from the output in the last layer of  encoder $\Phi(.)$.
Then \kmeans{}++~\cite{arthur2007km} is employed as an initialization of the seeding scheme for the following clustering process.
We denote the $k$-th cluster as $\mathcal{C}_k^{(t)} = \{ \mathrm{x}_i | c_i^{(t)} = k \}, k \in \{1, \ldots ,K\}$,
where $c_i^{(t)}$ is the cluster id for the instance  $\mathrm{x}_i$ at \al{} round $t$.
After obtaining $K$ clusters with the corresponding data $\mathcal{C}_k^{(t)}$,
we assign a pseudo label for each cluster.
First, the pseudo label for an individual instance $\mathrm{x}_i$ at round $t$ of \al{} is computed as: 
\begin{equation}
    \label{eq:pseudo_label}
    \widetilde{y}_i = \argmax_{j \in  \{1, \ldots ,Y\} }  [ \mathcal{M}(\mathrm{x}_i; \theta^{(t)} ) ]_j ,
\end{equation}
where $ \mathcal{M}(\mathrm{x}_i; \theta^{(t)} ) \in \mathbb{R}^Y $ is the probability distribution for  instance $\mathrm{x}_i$ over the $Y$ target classes,
 and  $[ \mathcal{M}(\mathrm{x}_i; \theta^{(t)} ) ]_j$ is the $j$-th entry denoting the probability 
 of $\mathrm{x}_i$ belonging to the target class $j$, inferenced by the current model.
Then the majority vote (the pseudo label of cluster $\mathcal{C}_k^{(t)}$) is derived as:
\begin{equation}
    \label{eq:pseudo_label_clu}
  y_{maj} = \argmax_j  (\sum_{i \in \mathcal{C}_{k}^{(t)}}{\mathbbm{1}\{\widetilde{y}_i=j}\})/ {|\mathcal{C}_{k}^{(t)}|}.
\end{equation}
The instances that are not predicted as $y_{maj}$ are defined as pseudo errors in the corresponding cluster $\mathcal{C}_k^{(t)}$.

\begin{algorithm*}[]
	\caption{ Round $t$ of \real \label{algo:real}}

    \DontPrintSemicolon
    \SetAlgoLined

	\KwIn{unlabeled pool $\Du$, budget for one iteration $b$, classification model $\mathcal{M}$,
              number of clusters $K$,  model's encoding part $\Phi(.)$}
        \KwOut{ sampled set $Q$ }
       
        $\mathcal{C}_k^{(t)} = \mathrm{KMeans}\big(\Phi(\Du) \big)$, $( k \in \{1, \ldots ,K\})$ \Comment{Clustering $\Du$}\;

	\For(\Comment{Process cluster $\mathcal{C}_k$}) {$ k \in \{1, \ldots ,K\}$}{ 
	    Run  Eq.~\ref{eq:pseudo_label} for cluster $\mathcal{C}_k$ to get the instance-level pseudo labels \; 

            Run  Eq.~\ref{eq:pseudo_label_clu} to find  the cluster-level pseudo label $y_{maj}$  for  $\mathcal{C}_k$ \;

            Init pseudo error set $\widetilde{E_k}$ in Eq.~\ref{eq:ep_clu} \;

	    Compute the error density $\epsilon_k$ for cluster $\mathcal{C}_k$ by Eq.~\ref{eq:ep_clu} \;
	}%
        
        Get the sampling budget $b_k$ based on error density for each cluster using Eq.~\ref{eq:budget_clu}\;
        
        \If { $\sum_k b_k < b $ }{
           $ \Delta $ = $b - \sum_k b_k$  \Comment{Budget residual}\;
        }
        
        $ b_k \mathrel{{+}{=}} 1,  \forall  k \in \Delta\text{-}\mathrm{argmax}_k( b_k) ~\text{and}~ b_k >0 $  \label{lst:line:blah2}  
          \Comment{Allocate residual to top-$\Delta$ largest $b_k$}\ \;
        
        $Q= \emptyset $  \Comment{Init the sample set}\;

        \For {$ k \in \{1, \ldots ,K\} $}{    
            Random sample $\min( |\widetilde{E_k}|, b_k )$ instances from $\widetilde{E_k}$  into $Q$  \label{lst:line:minekbk} \;
	}%

        \If { $ |Q| < b $ }{
           $Q = Q ~\cup $   $\{( b-|Q|)$  instances from $\Du$  with top $\epsilon(.)$ scores (Eq.~\ref{eq:ep_ins}) and  not in $Q \}$    \label{lst:line:complement} \;
        }

\end{algorithm*}

\subsection{Adaptive Sampling of Representative Errors}

For each round of active learning, assuming that the labeling budget is $b$, we need to decide how we should select the $b$ samples from the unlabeled pseudo errors.
To ensure the representativeness of selected samples,
we allocate the sampling budget $b$ to each cluster according to the density of pseudo errors in the cluster, i.e., the percentage of the pseudo errors within a cluster over the total number of pseudo errors in the whole unlabeled data pool.
A larger sampling budget will be allocated to the cluster with a higher pseudo error density.
The density of pseudo errors $\epsilon_k$ for cluster $\mathcal{C}_{k}^{(t)}$ is defined as:
\begin{equation}
    \label{eq:ep_clu}
       \epsilon_k = \sum_{\mathrm{x}_e \in \widetilde E_k}  \epsilon(\mathrm{x}_e), 
\end{equation}
where $\widetilde E_k =\{ \mathrm{x}_e | \mathrm{x}_e \in  \mathcal{C}_{k}^{(t)} ~ \text{and} ~ \widetilde{y}_e \neq   y_{maj} \} $ is the pseudo error set  in the $k$-th cluster, 
and $\epsilon(\mathrm{x}_e)$ is one pseudo error $\mathrm{x}_e$'s contribution to the cluster-level error density:
\begin{equation}
    \label{eq:ep_ins}
         \epsilon(\mathrm{x}_e) =1- [ \mathcal{M}(\mathrm{x}_e; \theta^{(t)} ) ]_{maj} .
\end{equation}
The sampling budget $b_k$ for  the $k$-th cluster is then normalized as:
\begin{equation}
    \label{eq:budget_clu}
       b_k = \left\lfloor b \dfrac{ \epsilon_k }
                     { \sum_i \epsilon_i } \right\rfloor , \forall  k \in \{1 \ldots K\} .
\end{equation}
Apart from selecting pseudo errors, we also try to select errors near the classification decision boundary
by emphasizing clusters with  denser pseudo errors.
The empirical evidences in \S\ref{sec:experiment_results} also show  that our adaptive budget allocation is able to pick more \emph{representative} pseudo errors along the decision boundary.

In real-world applications of \al{},
it is possible that there may not be enough pseudo errors to be sampled in a cluster (i.e., $ |E_k| < b_k$).
For instance, when the model is already well-trained via active learning, most of the data instances will be correctly classified.
In those cases, we complement the sampled set $Q$ by instances with a higher erroneous probability within all the unlabeled pool $\Du$  (Line~\ref{lst:line:complement} in Algorithm~\ref{algo:real}), which are illustrated as blue instances (\toyPseTriBlue{} \toyPseCirBlue{}) in Fig.~\ref{fig:real_toy}.
The complexity of \real{} consists of two parts: the inference time $O(|\Du|)$ and the time for \kmeans{}
clustering $O(dK|\Du|)$, where $d$ is the encoder feature dimension $|\Phi(.)|$.
\kmeans{} implemented in faiss~\cite{johnson2019faiss} costs only 2 or 3 seconds even for large datasets such as \ag{} and \pub{} in \S\ref{sec:experiment_setup}.

\section{Experimental Setup}
\label{sec:experiment_setup}

\subsection{Datasets}

Following prior research~\cite{yu2022actune,margatina2021active,yuan2020alps},
We conduct experiments on five text classification datasets from different application domains, i.e.,
\sst{}~\cite{socher-etal-2013-recursive},
\ag{}~\cite{zhang2015character},
\pub{}~\cite{dernoncourt-lee-2017-pubmed},
\snips{}~\cite{coucke2018snips},
and \stov{}~\cite{xu2015stackoverflow}.
Table~\ref{tab:datasets} shows their detailed statistics.
Due to the limited computational resources,
we follow the  prior study~\cite{yu2022actune} and take a subset of the original training set and validation set if they are too large.
Specifically, we randomly sample 20K$\times Y$ instances form each training set if its size exceeds 20K$\times Y$,
where $Y$ is the number of target classes.
We also keep the size of validation set no more than 3K to speed up the validation process.
\begin{table*}[htb]
\centering
\vspace{-0.4cm}
\setlength{\tabcolsep}{6pt} %
\renewcommand{\arraystretch}{1.1} %
\caption{Dataset statistics.}
\label{tab:datasets}
\small
\begin{tabular}{lccccccc}
\toprule
\textsc{dataset} &  \textsc{label type}   & \#\textsc{train} & \#\textsc{val}  & \#\textsc{test} & \#\textsc{classes}\\
\midrule 
\sst{} & Sentiment        & 40K  &  3K &  1.8K &  2 \\
\ag{} &  News Topic      &  80K  &  3K &  7.6K &  4 \\
\pub{} &  Medical Abstract  &  100K  &  3K &  30.1K &  5 \\
\snips{}  &  Intent   &  13K &  0.7K &  0.7K &  7 \\
\stov{}  &  Question   & 8.0K &  1K &  1K &  10 \\
\bottomrule
\vspace{-0.6cm}
\end{tabular}
\end{table*}

\subsection{Baselines \& Implementation Details}
We compare \real{} against 8 baselines:
    (1) \entropy{} selects instances with the most even distribution of prediction probability~\cite{lewis1995sequential};
    (2)  \plmkm{} \cite{yuan2020alps} is a diversity-based baseline which selects $b$ instances closest to \kmeans{} centers of the \cls{} token embeddings;
    (3)  \badge{}~\cite{ash2020deep} transforms data into gradient embeddings that encode the model confidence and then  use \kmeans{}++ to select;
    (4)  \bald{}~\cite{gal2017deep} defines uncertainty as the mutual information among different versions (via multiple MC dropouts~\cite{gal2015bayesian}) of the model's predictions;
    (5)  \alps{}~\cite{yuan2020alps} selects  by  masked language modeling loss in \plm{};
    (6)  \call{}~\cite{margatina2021active} tries to sample the most contrastive instances along the classification decision boundary;
    (7)  \actune{}~\cite{yu2022actune} selects the unlabeled samples with high uncertainty for active annotation and those with low uncertainty for semi-supervised self-training by weighted \kmeans{};
    Since semi-supervised self-training is out of the scope of our current work, we remove the self training part from \actune{} for a fair comparison.
    (8)  \random{} uniformly samples data from the unlabeled  pool $\Du$.

For the text classification model $\mathcal{M}$, we follow the prior study~\cite{yu2022actune} and use \roberta{}-\textsc{base}~\cite{liu2019roberta} implemented in the HuggingFace library~\cite{wolf-etal-2020-transformers} for our experiments.
We train the model on the initial warm-up labeled set for 10 epochs,
and continually train the model for 4 epochs after each round of active sampling to avoid overfitting.
We evaluate the model 4 times per training epoch on the validation set and keep the best version.
At the end of training, we test the previously-saved best model on the hold-out test set.
We choose the best hyperparameters for baselines as indicated in their original papers.
We set the \al{} rounds to be 8 for all the 9 \al{} methods.
Following \cite{yuan2020alps,margatina2021active}, all of our methods and
baselines are run with 4 different random seed and the result is based on the average performance
on them.
This creates 5 (datasets) × 4 (random seeds) ×
13 (8 baselines + 1 \real{} + 4 \real{} variants) ×  8 (rounds) = 2080 experiments,
 which is almost the limit of our computational resources.
More details on the experiment setup can be found in our code repository.
\section{Results}
\label{sec:experiment_results}
In the experimental study,
 we try to answer the following research questions:
\begin{itemize}
    \item \textbf{RQ1.} \emph{Classification Performance}: How is the classification performance of \real{} compared to baselines? (\S\ref{sec:overall-perf})
    \item \textbf{RQ2.} \emph{Representative errors}:  What are the characteristics of the samples, e.g., error rate and  representativeness? (\S\ref{sec:analysis})
    \item  \textbf{RQ3.} \emph{Ablation \& hyperparameter}: What is the performance of different design variants of \real{}? How robust is \real{} under different hyperparameter settings? (\S\ref{sec:ablation})
\end{itemize}

\begin{table*}[ht]
\vspace{-0.3cm}
\setlength{\tabcolsep}{4pt} %
\caption{Mean accuracy (top half) and mean F1-macro (bottom half).}
\label{tab:overall_acc}
\centering
\begin{tabular}{lccccccccc}
\toprule
\textsc{dataset} & \scriptsize \entropy{} &  \scriptsize  \plmkm{} & \scriptsize  \bald{} &  \scriptsize  \badge{} & \scriptsize  \alps{} & \scriptsize  \call{} & \scriptsize \actune{} & \scriptsize  \random{} &   \real{} \\
\midrule
\sst{}  &   91.95 &  90.74 &  90.23 &  90.90 &  90.88 &  91.49 &  91.82 &  89.91 &  \textbf{92.41} \\
\ag{} &   90.34 &  90.19 &  89.73 &  90.11 &  89.70 &  90.65 &  90.57 &  89.23 &  \textbf{91.08} \\
\pub{} &   80.99 &  80.90 &  79.06 &  81.28 &  79.93 &  81.26 &  81.53 &  80.60 &  \textbf{82.17} \\
\snips{}  &   95.99 &  95.51 &  95.26 &  95.52 &  94.93 &  96.07 &  96.16 &  94.81 &  \textbf{96.63} \\
\stov{}   &   86.56 &  85.83 &  85.40 &  86.39 &  85.62 &  86.67 &  86.27 &  84.82 &  \textbf{87.37} \\
\hline
\sst{}  &   91.94 &  90.47 &  90.22 &  90.90 &  90.95 &  91.48 &  91.82 &  89.90 &  \textbf{92.41} \\
\ag{} &   90.35 &  89.84 &  89.66 &  90.12 &  89.60 &  90.26 &  90.66 &  89.31 &  \textbf{90.96} \\
\pub{} &   73.92 &  74.21 &  71.54 &  74.82 &  73.48 &  74.99 &  75.30 &  73.90 &  \textbf{75.78} \\
\snips{}  &   96.06 &  95.60 &  95.35 &  95.58 &  94.82 &  96.13 &  96.22 &  94.86 &  \textbf{96.69} \\
\stov{}   &   86.76 &  85.96 &  85.46 &  86.53 &  85.72 &  86.83 &  86.41 &  84.99 &  \textbf{87.53} \\
\bottomrule
\vspace{-0.3cm}
\end{tabular}
\end{table*}
\subsection{Classification Performance}
\label{sec:overall-perf}

For \textbf{RQ1}, 
we compare the classification performance of \real{} against state-of-the-art baselines.
Following the existing work of text classification~\cite{lai2015recurrent}, we use two criteria: accuracy and F1-macro
to measure the model performances.
Accuracy is the fraction of predictions our model got right.
F1-macro is the average of accuracy independently measured for each class (i.e., treating different classes equally).

\begin{figure}[ht]
     \begin{subfigure}[t]{0.32\textwidth}
         \centering
         \includegraphics[width=\textwidth]{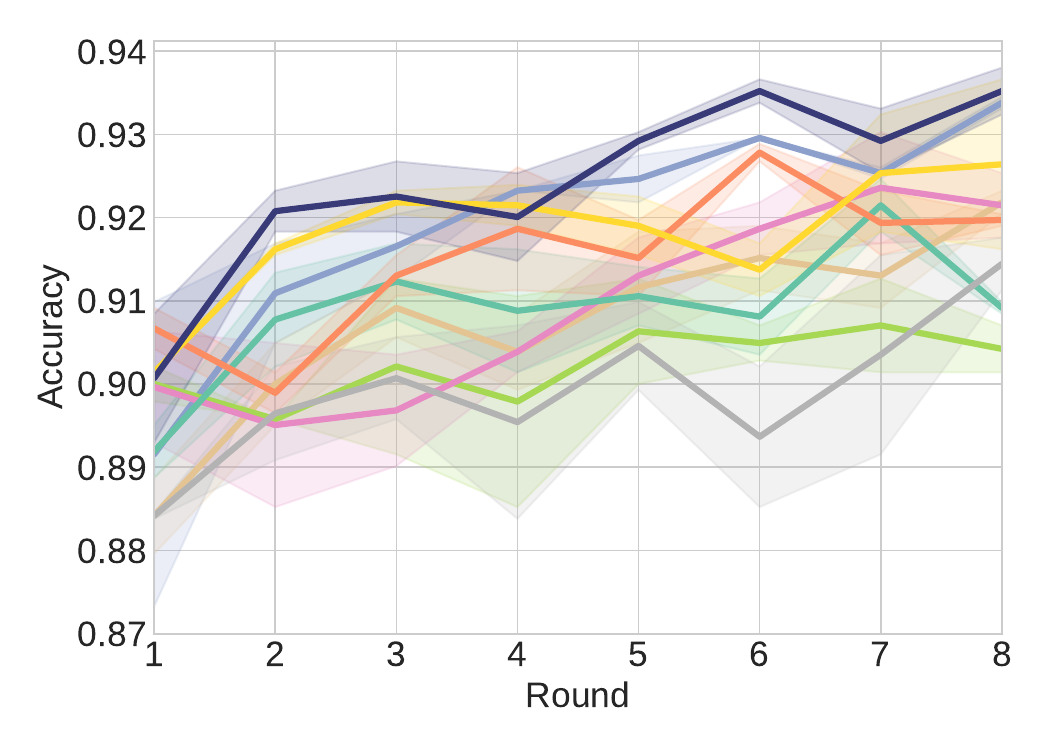}
         \caption{\sst{}}
         \label{fig:oacc_sst}
     \end{subfigure}
     \hfill
     \begin{subfigure}[t]{0.32\textwidth}
         \centering
         \includegraphics[width=\textwidth]{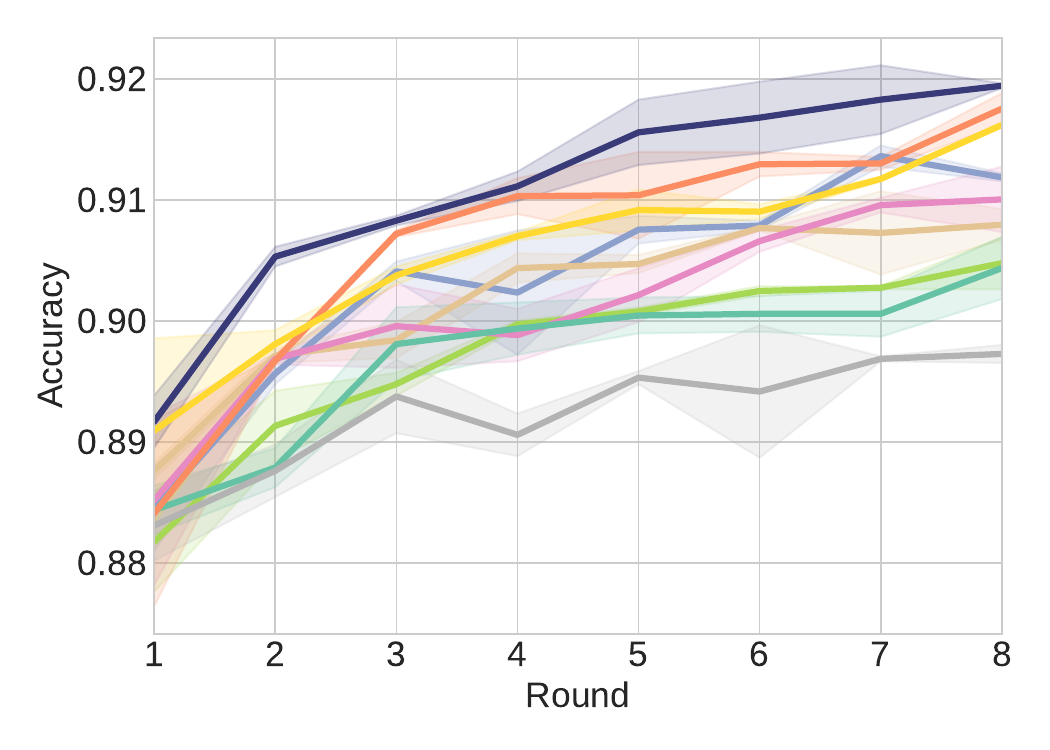}
         \caption{\ag{}}
         \label{fig:oacc_agnews}
     \end{subfigure}
     \hfill
     \begin{subfigure}[t]{0.32\textwidth}
         \centering
         \includegraphics[width=\textwidth]{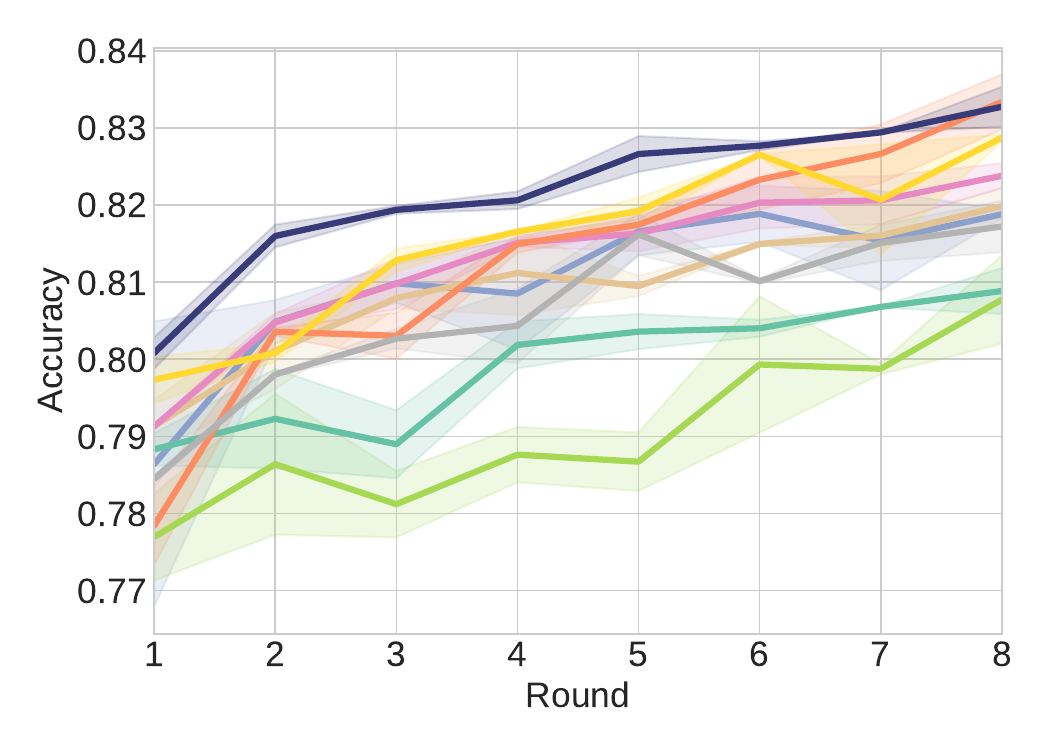}
         \caption{\pub{}}
         \label{fig:oacc_pubmed}
     \end{subfigure}
     \hfill %
     \begin{subfigure}[b]{0.32\textwidth}
         \centering
         \includegraphics[width=\textwidth]{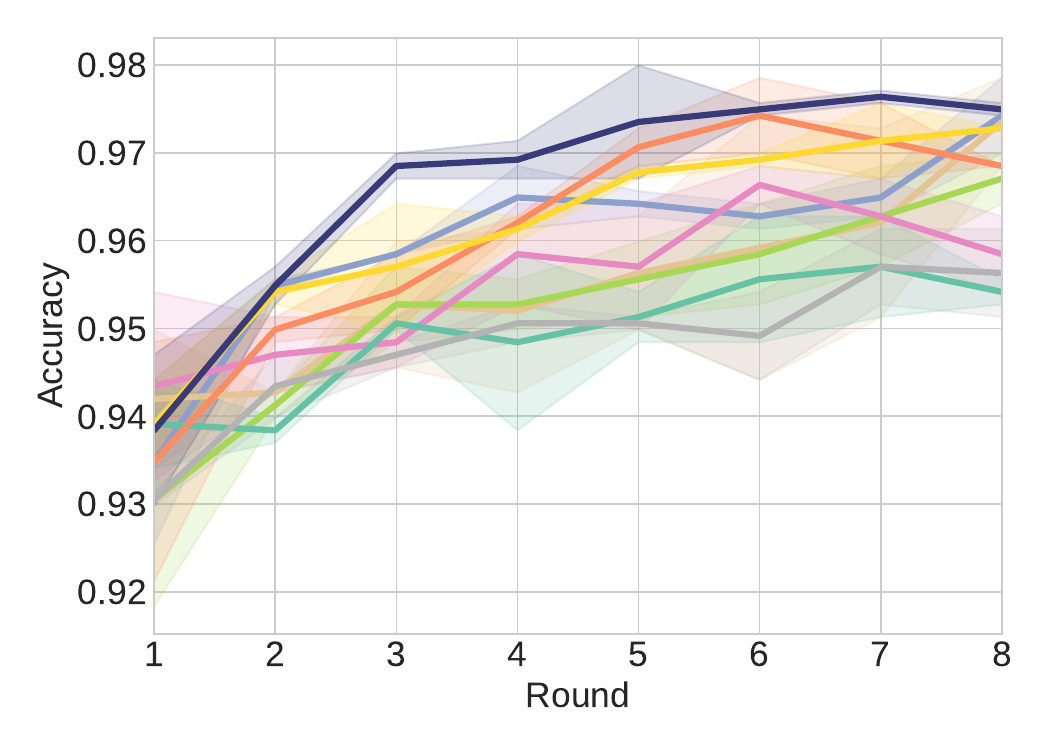}
         \caption{\snips{}}
         \label{fig:oacc_snips}
     \end{subfigure}
     \hfill
     \begin{subfigure}[b]{0.32\textwidth}
         \centering
         \includegraphics[width=\textwidth]{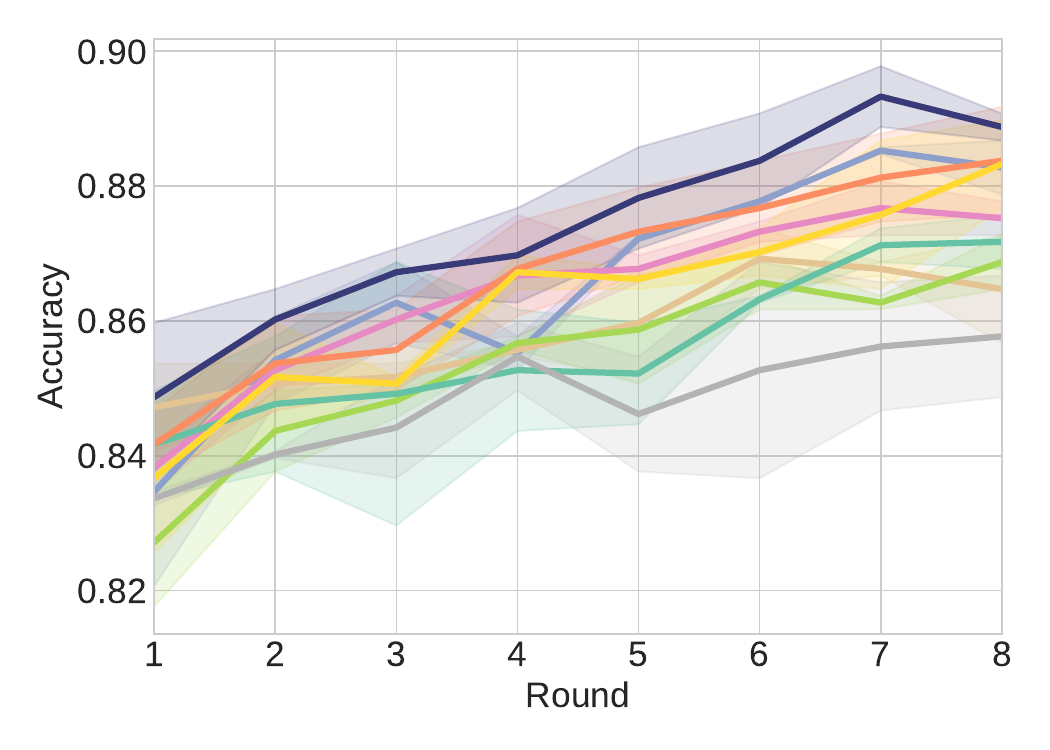}
         \caption{\stov{}}
         \label{fig:oacc_trec}
     \end{subfigure}
     \hfill
     \begin{subfigure}[b]{0.32\textwidth}
         \centering
         \includegraphics[width=\textwidth]{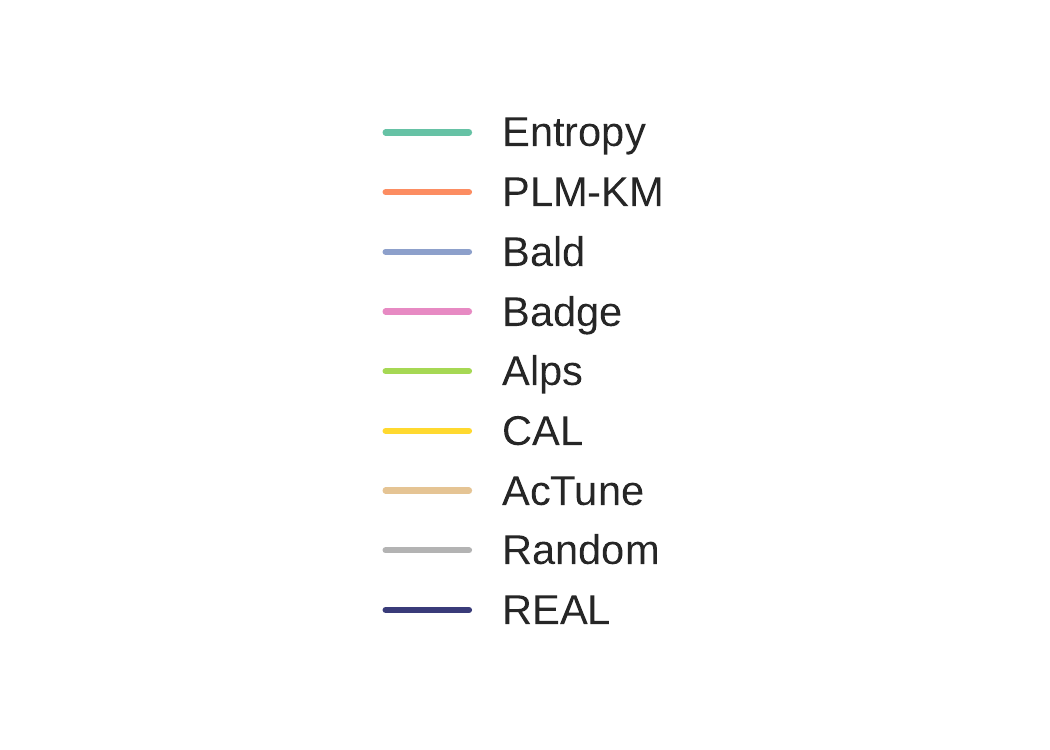}
         \caption{Legend}
         \label{fig:oacc_ledgend}
     \end{subfigure}
\caption{The accuracy of different active learning rounds on each dataset. The height of the shadow area denotes the std of accuracy.} \label{fig:overall_acc}
\vspace{-0.3cm}
\end{figure}

Table \ref{tab:overall_acc} shows
the average accuracy and F1-macro scores
of different \al{} strategies on all the datasets.
The detailed accuracy for each \al{} round is shown in Fig.~\ref{fig:overall_acc}.
\real{} outperforms all the baselines by 0.43\% -- 0.70\%
performance gain
w.r.t. the mean accuracy of all the eight \al{} rounds.
The two most recent baselines, \call{} and \actune{} rank in the second or third places in most cases, which is consistent with the reports in their original papers.
\entropy{} is also a strong baseline and performs better than \plmkm{} and \alps{}.
The relatively good performance of \entropy{} 
can be explained by pretrained language model's good uncertainty estimations~\cite{desai2020calibration}.
It stands out in \sst{} dataset, probably because it is relatively easy to pick samples around the decision boundary for the binary classification task. As shown in Fig.~\ref{fig:overall_acc},
\pub{} is a difficult dataset to learn.
The model's  test accuracy on it is less than 85\%  even with the full training set.
It is probably because the professional medical text is rarely seen in \roberta{} and the label distribution of \pub{} is a skewed.
\entropy{} and \bald{} perform very badly on \pub{}, since they heavily rely on the distribution of the prediction probability. Compared to other baseline methods, \real{} has a clear advantage on \pub{}.

\subsection{Representative Errors}
\label{sec:analysis}
We address \textbf{RQ2} by investigating whether \real{} can sample \emph{representative errors} and comparing it with those baselines.
Specifically, 
we evaluate the capability of \real{} in selecting representative error samples from the following two perspectives:
\begin{itemize}
    \item The error rate and initial training loss of samples (Table~\ref{tab:error-rate} and Fig.~\ref{fig:celoss-errd});
    \item The distribution divergence between samples and boundary errors (Fig.~\ref{fig:celoss-errd}).
\end{itemize}

\subsubsection{Error Rate and Initial Training Loss of Samples}

Table~\ref{tab:error-rate} shows the mean error rates and initial training loss (for all \al{} rounds) of samples $Q$ for different \al{} strategies across all the datasets.
The error rate $\varepsilon(Q)$ is the proportion of wrongly-predicted instances in $Q$ by comparing the model prediction with the ground truth label for each instance.
It is inappropriate to directly compare the error rates $\varepsilon(Q)$ of different \al{} strategies,
because the error rates $\varepsilon(\Du)$ in the whole unlabeled pool $\Du$ are different.
It is more easier to achieve a high sampling error rate $\varepsilon(Q)$ given a high background error rate $\varepsilon(\Du)$.
Therefore, we compares the \lift{} of sampling error rate, which is defined as $\varepsilon(Q) / \varepsilon(\Du)$,
which implies how effective does an \al{} strategy select errors, compared with random selection.
Another metric is the average cross entropy loss $\ell_0$ of samples $Q$ in the first training step
, which is a more fine-grained version of error rate.
Many previous research work \cite{yuan2021multiple,luo2021loss,wan2021unsupervised,huang2021semi}
have already validated that samples with higher loss are usually more informative to the model.

Table~\ref{tab:error-rate} shows 
\real{} usually has a large \lift{} of sampling error rate, second only to \entropy{}, despite the fact that 
the unlabeled pool error rate $\varepsilon(\Du)$ of \real{} is the  lowest.
The large \lift{} of sampling error rate implies \real{} successfully identifies the errors in $\Du$.
It is worth mentioning that  $\varepsilon(\Du)$ can serve as test set because we don't use $\Du$ in the previous \al{} rounds 
for training or validation.
\real{} has the lowest error rate on 4 out of 5 datasets, which means the highest accuracy when testing on  $\Du$.
The initial training loss of \real{}'s samples is also the largest in the last four datasets.
Fig.~\ref{fig:celoss-errd} shows more detailed loss distribution for each \al{} round.
\begin{table*}[!htbp]
\caption{  \small  Sampling error rate $\varepsilon(.)$, \lift{}, and the average first step training  loss  $\ell_0$.}
\label{tab:error-rate}
\centering
\begin{tabular}{llrrrrrrr}
\toprule
\textsc{dataset}  &  \scriptsize \textsc{metric} &  \scriptsize \entropy &  \scriptsize \plmkm &  \scriptsize \badge &  \scriptsize   \call{}~~ &  \scriptsize \actune &  \scriptsize \random &   \small  \real \\
\midrule
\multirow{4}{*}{ \sst{} } & $\varepsilon(Q)$ &   \textbf{0.4959} & 0.1841 & 0.2308 &  0.4821 &  0.4334 &  0.1284 &  0.4739 \\
   &    $\varepsilon(\Du)$ &   \textbf{0.1194} & 0.1251 & 0.1259 &  0.1215 &  0.1170 &  0.1338 &  0.1212 \\
   &    \lift{} &   \textbf{4.1530} & 1.4713 & 1.8325 &  3.9670 &  3.7055 &  0.9596 &  3.9113 \\
   &     $\ell_0$  &   0.6984 & 0.8100 & \textbf{1.0538} &  0.6915 &  0.8526 &  0.6660 &  0.9938 \\
\hline
\multirow{4}{*}{ \ag{} } & $\varepsilon(Q)$ &   \textbf{0.6092} & 0.1904 & 0.2246 &  0.5637 &  0.5325 &  0.1142 &  0.5537 \\
 &    $\varepsilon(\Du)$ &   0.1009 & 0.1039 & 0.1041 &  0.0995 &  0.0991 &  0.1115 &  \textbf{0.0959} \\
 &    \lift{} &   \textbf{6.0377} & 1.8320 & 2.1576 &  5.6667 &  5.3730 &  1.0239 &  5.7737 \\
 &     $\ell_0$  &   1.2504 & 0.8597 & 0.9477 &  1.0926 &  1.3009 &  0.5707 &  \textbf{1.3636} \\
\hline
\multirow{4}{*}{ \pub{} } & $\varepsilon(Q)$ &   \textbf{0.6701} & 0.3164 & 0.3634 &  0.6103 &  0.6231 &  0.1987 &  0.6046 \\
 &    $\varepsilon(\Du)$ &   0.1943 & 0.1971 & 0.1928 &  0.1941 &  0.1907 &  0.1998 &  \textbf{0.1858} \\
 &    \lift{} &   \textbf{3.4487} & 1.6048 & 1.8845 &  3.1452 &  3.2670 &  0.9943 &  3.2531 \\
 &     $\ell_0$  &   1.5117 & 1.3533 & 1.6009 &  1.2871 &  1.4494 &  1.0222 &  \textbf{1.7040} \\
\hline
\multirow{4}{*}{ \snips{} }  & $\varepsilon(Q)$ &   0.4107 & 0.1226 & 0.1120 &  \textbf{0.4237} &  0.2963 &  0.0276 &  0.4002 \\
  &    $\varepsilon(\Du)$ &   0.0268 & 0.0337 & 0.0308 &  0.0280 &  0.0265 &  0.0393 &  \textbf{0.0231} \\
  &    \lift{} &  15.3183 & 3.6410 & 3.6338 & 15.1568 & 11.1895 &  0.7023 & \textbf{17.2902} \\
  &     $\ell_0$  &   \textbf{1.0176} & 0.5209 & 0.5080 &  1.0470 &  0.9491 &  0.1842 &  0.9356 \\
\hline
\multirow{4}{*}{ \stov{} }  & $\varepsilon(Q)$ &   \textbf{0.7328} &  0.2536 & 0.3506 &  0.6904 &  0.6659 &  0.1307 &  0.7162 \\
   &     $\varepsilon(\Du)$ &   0.1048 &  0.1263 & 0.1209 &  0.1094 &  0.1101 &  0.1386 &  \textbf{0.1045} \\
   &    \lift{} &   \textbf{6.9934} &  2.0079 & 2.8994 &  6.3114 &  6.0509 &  0.9435 &  6.8548 \\
   &     $\ell_0$  &   \textbf{2.1434} &  1.0260 & 1.3874 &  2.0255 &  2.0062 &  0.6331 &  2.1131 \\
\bottomrule
\vspace{-0.3cm}
\end{tabular}
\end{table*}

\subsubsection{Representativeness}

\begin{figure}[t]
     \begin{subfigure}[t]{0.32\textwidth}
         \centering
         \includegraphics[width=\textwidth]{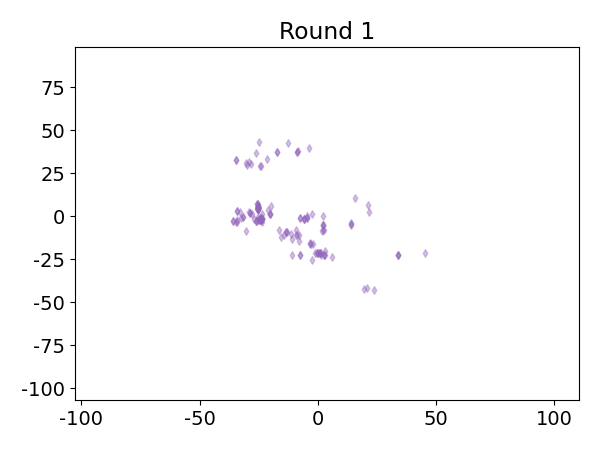}
         \caption{ \entropy{} }
         \label{fig:tsne_err-r1_entropy}
     \end{subfigure}
     \hfill
     \begin{subfigure}[t]{0.32\textwidth}
         \centering
         \includegraphics[width=\textwidth]{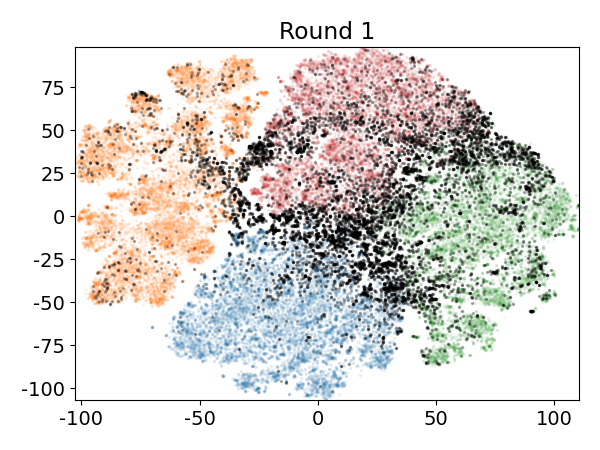}
         \caption{ Errors (in black)}
         \label{fig:tsne_err-r1_error}
     \end{subfigure}
     \hfill
     \begin{subfigure}[t]{0.32\textwidth}
         \centering
         \includegraphics[width=\textwidth]{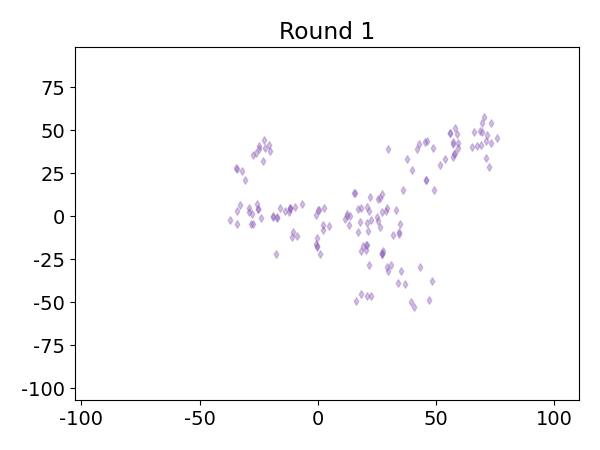}
         \caption{\real{}}
         \label{fig:tsne_err-r1_real}
     \end{subfigure}
     \vfill
     \begin{subfigure}[b]{0.32\textwidth}
         \centering
         \includegraphics[width=\textwidth]{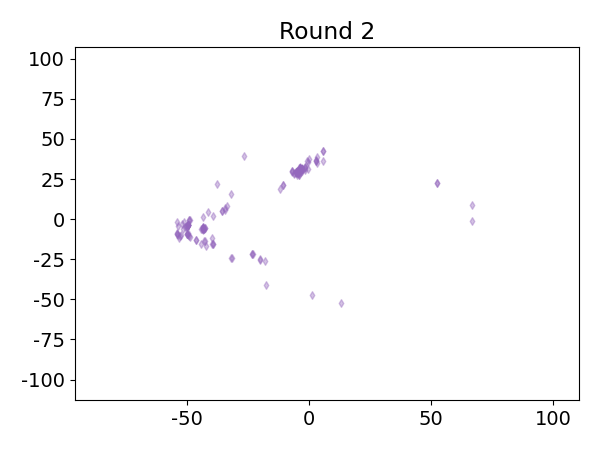}
         \caption{ \entropy{} }
         \label{fig:tsne_err-r2_entropy}
     \end{subfigure}
     \hfill
     \begin{subfigure}[b]{0.32\textwidth}
         \centering
         \includegraphics[width=\textwidth]{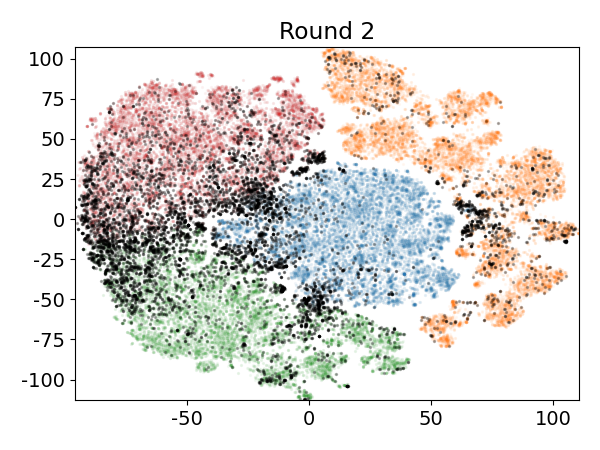}
         \caption{Errors (in black)}
         \label{fig:tsne_err-r2_error}
     \end{subfigure}
     \hfill
     \begin{subfigure}[b]{0.32\textwidth}
         \centering
         \includegraphics[width=\textwidth]{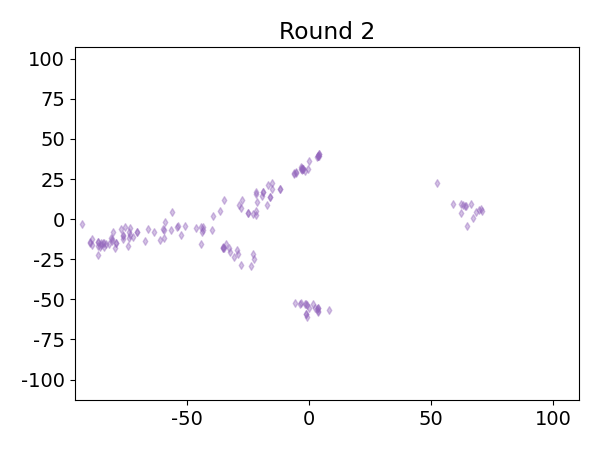}
         \caption{\real{}}
         \label{fig:tsne_err-r2_real}
     \end{subfigure}
\caption{The t-SNE based visualization for the sample/error distribution.
 Visualizations in the first row are from the first round active selection on \ag{} and the second row  from the second round.
 Two subfigures in the middle (\ref{fig:tsne_err-r1_error} and \ref{fig:tsne_err-r2_error}) are all instances in $\Du$. The four colors in \ref{fig:tsne_err-r1_error} and \ref{fig:tsne_err-r2_error} indicate different correctly predicted categories, except black. Black dots indicate the ground truth errors.  Purple dots in the left side two visualizations are \entropy{}'s selections and the right side are \real{}'s selections. }
 \vspace{-0.3cm}
 \label{fig:tsne-err}
\end{figure}

We investigate how \real{}'s  samples align with ground-truth errors on decision boundary compared to  baselines.
Based on theoretical studies on margin theory for active learning \cite{balcan2007margin},
selecting instances close to the decision boundary can significantly reduce the number of annotations required \cite{margatina2021active,ru2020active,zhu2017generative,huijser2017active}.
Though identifying the precise decision boundaries for deep neural networks is intractable \cite{ducoffe2018adversarial}, we use the basic grid statistics on t-SNE embeddings as an empirical solution.
Specifically, we apply t-SNE to the \cls{} token embeddings of $ \Phi( \Du)$ 
and project the original token embeddings of 768 dimensions to 2D plane,
as shown in Fig.~\ref{fig:tsne-err}.
Then, we split the bounding box of $\Du$ on 2-D plane into $50 \times 50$ uniform grids, where
 $g_i$ denotes the number of instances that fall into the $i$-th grid.
\begin{figure}[htbp]
     \begin{subfigure}[t]{0.5\textwidth}
         \centering
         \includegraphics[width=\textwidth]{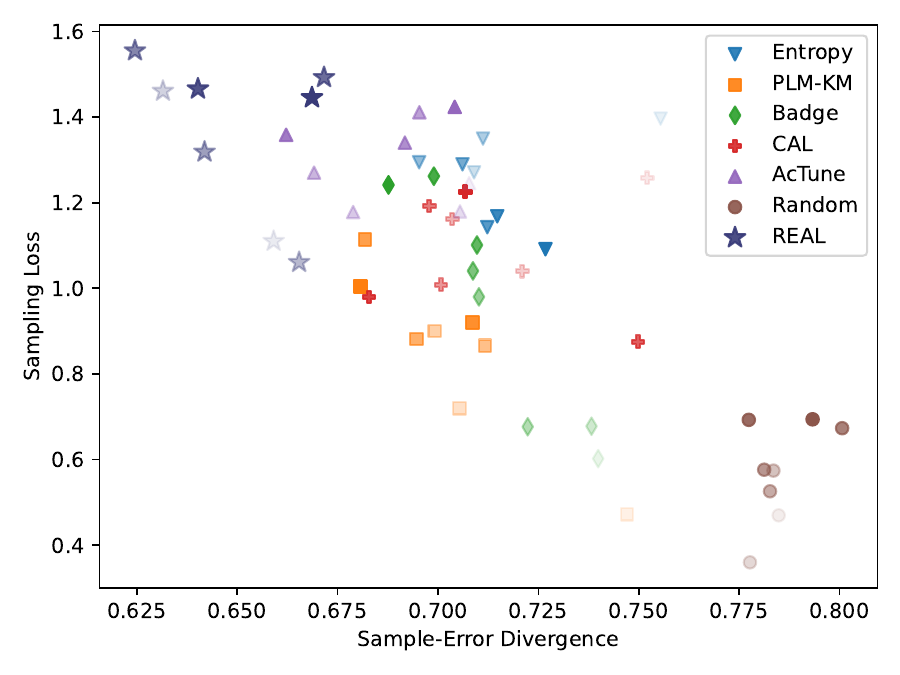}
         \caption{\ag{}}
         \label{fig:celoss-errd-agnews}
     \end{subfigure}
     \hfill
     \begin{subfigure}[t]{0.5\textwidth}
         \centering
         \includegraphics[width=\textwidth]{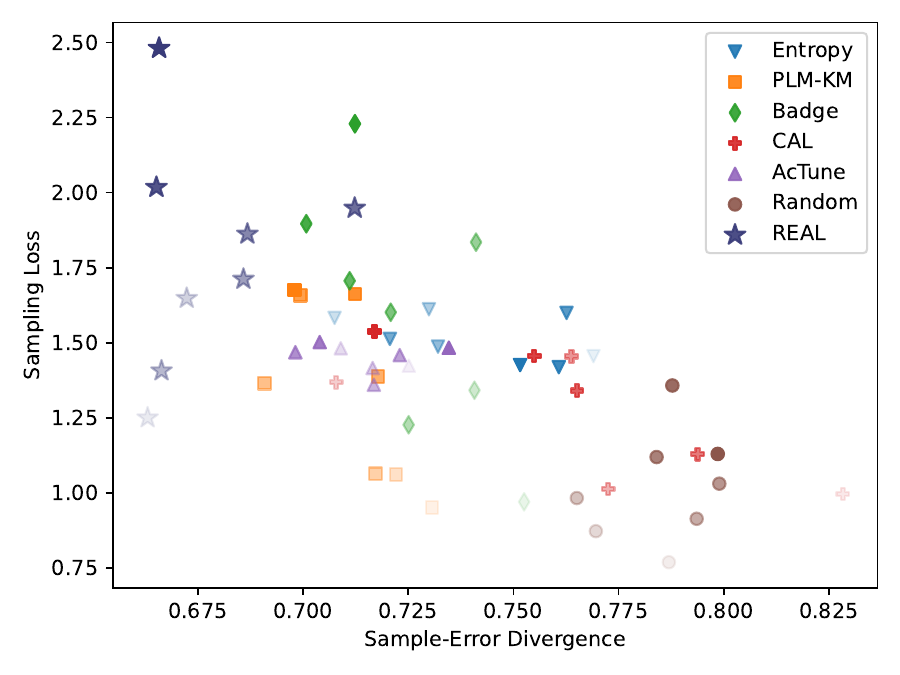}
         \caption{\pub{}}
         \label{fig:celoss-errd-pubmed}
     \end{subfigure}
\caption{ \small Loss v.s. sample-error divergence. Each dot represents a sample set $Q$ from one \al{} round.
The dot shape and color hue indicate \al{} strategy.
The dot transparency indicates different \al{} rounds.
More transparent dot comes from earlier \al{} round, and darker dot comes from later \al{} round.
\real{} clearly shows a lower divergence to the ground truth errors on the decision  boundary, and a larger sampling loss.}
\vspace{-0.3cm}
\label{fig:celoss-errd}
\end{figure}

We keep only the ground-truth errors within the decision boundaries.
Our intuition of decision boundary is where instances have similar representations but different predictions~\cite{balcan2007margin}.
We hypothesize that the pseudo errors selected by \real{} is near to the model’s decision boundary because 
1) instances in the same cluster have similar representations;
and 2) pseudo errors have different predictions than the majority in a cluster.
To verify this hypothesis, we compute the distribution entropy of ground-truth labels for each grid, and reserve only the top $0.15$ grids with high entropy values.
For example, in binary classification, a ground-truth label distribution $[98,100]$ for grid $g_1$ means there are 98 instances belonging to class-0 and 100 instances belonging to class-1 in $g_1$.
Suppose another grid  $g_2$ has ground-truth label distribution $[180, 10]$.
Then $g_1$ will be closer to the decision boundary than $g_2$ because $g_1$ has a more even distribution, thus a larger entropy.
We call grids with top $15\%$   high entropy values as boundary grids.
 $g_i^\varepsilon$ denotes the number of ground-truth errors fall into $i$-th boundary grid.
We compare the Jensen-Shannon divergence ($\mathrm{JSD}$)  between boundary grids set
$\{g_i^\varepsilon\}_{i=1}^{m}$ ($m$ is the  number of boundary grids)
and $\{s_i\}_{i=1}^{m}$, where $s_i$ is the number of sampled instances in the $i$-th grid.
$\mathrm{JSD}$ extends KL divergence ($\mathrm{KLD}$) to derive a symmetric distance measure between two probability distributions $P_1$ and $P_2$: $ \mathrm{JSD}(P_1 \| P_2)=\frac{1}{2} \mathrm{KLD}(P_1 \| M)+\frac{1}{2}\mathrm{KLD}(P_2 \| M)$,
where $M=\frac{1}{2}(P_1+P_2)$.
Together with previously introduced sampling loss,
we plot the divergence in Fig.~\ref{fig:celoss-errd} for the  two largest datasets \ag{} and \pub{}.
\real{} clearly has the lowest divergence,
which means our samples' distribution aligns well with the ground-truth errors on the boundary.

Besides the lowest divergence, \real{} also has the largest sampling loss.
As shown in Fig.~\ref{fig:celoss-errd}, \real{}'s samples clearly distribute in the upper left corner.
In contrast, \random{}'s samples lie in the lower right corner. 
The least sampling loss and largest divergence from boundary errors may be the reason why \random{} fails.
To our surprise, darker dots usually appear in higher positions, in Fig.~\ref{fig:celoss-errd}, which means samples in later \al{} rounds provide larger loss.
The reason may be that the model in later \al{} rounds is stable and confident in its predictions,
thus introducing new samples will cause a larger loss.

Fig.~\ref{fig:tsne-err} provides the case studies of our samples against \entropy{} on \ag{}.
We can see that most errors distribute near the decision boundaries.
\entropy{} tends to miss some decision  boundary areas and is lack of diversity.
\real{} matches the boundary errors better.

\subsection{Ablation and Hyperparameter Study}
\label{sec:ablation}
\begin{figure}[t]
     \begin{subfigure}[t]{0.32\textwidth}
         \centering
         \includegraphics[width=\textwidth]{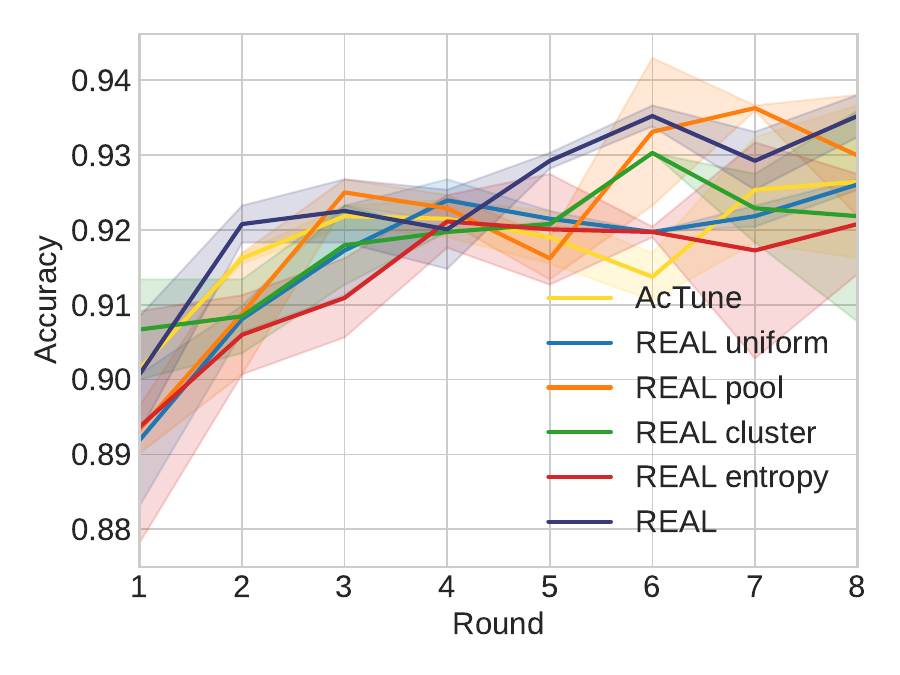}
         \caption{\sst{}}
         \label{fig:abl_sst}
     \end{subfigure}
     \hfill
     \begin{subfigure}[t]{0.32\textwidth}
         \centering
         \includegraphics[width=\textwidth]{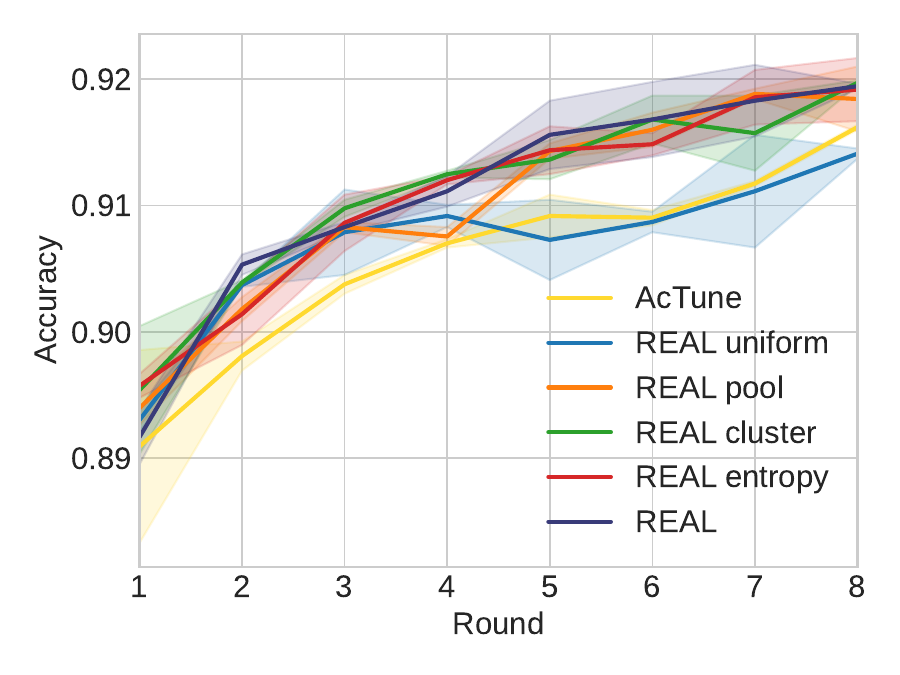}
         \caption{\ag{}}
         \label{fig:abl_agnews}
     \end{subfigure}
     \hfill
     \begin{subfigure}[t]{0.32\textwidth}
         \centering
         \includegraphics[width=\textwidth]{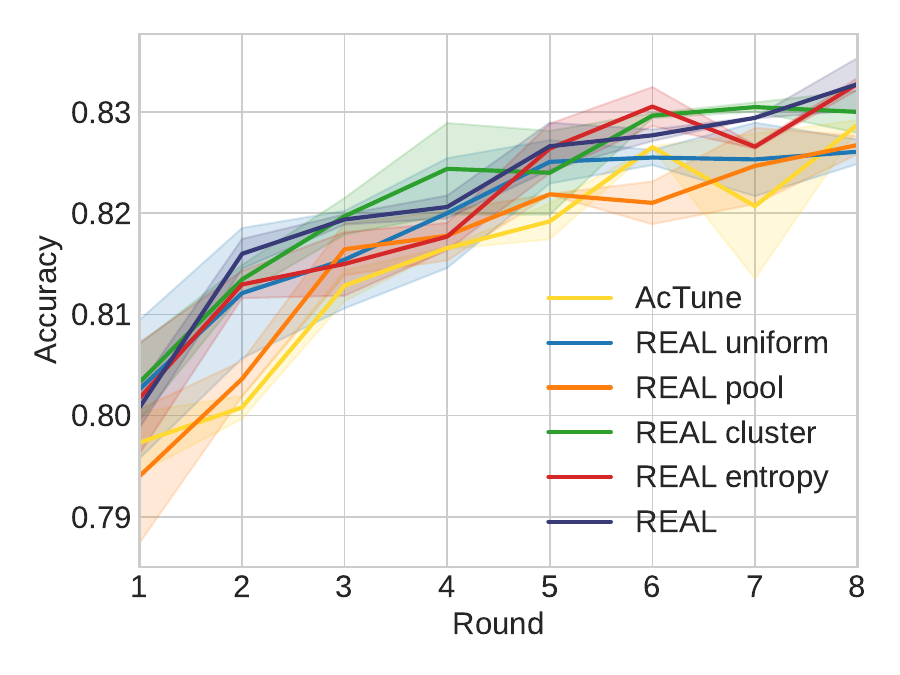}
         \caption{\pub{}}
         \label{fig:abl_pubmed}
     \end{subfigure}
\caption{Ablation study on different variants of \real{}.} 
\label{fig:ablation}
\end{figure}

\begin{figure}[ht]
     \centering
     \begin{subfigure}[t]{0.45\textwidth}
         \centering
         \includegraphics[width=\textwidth]{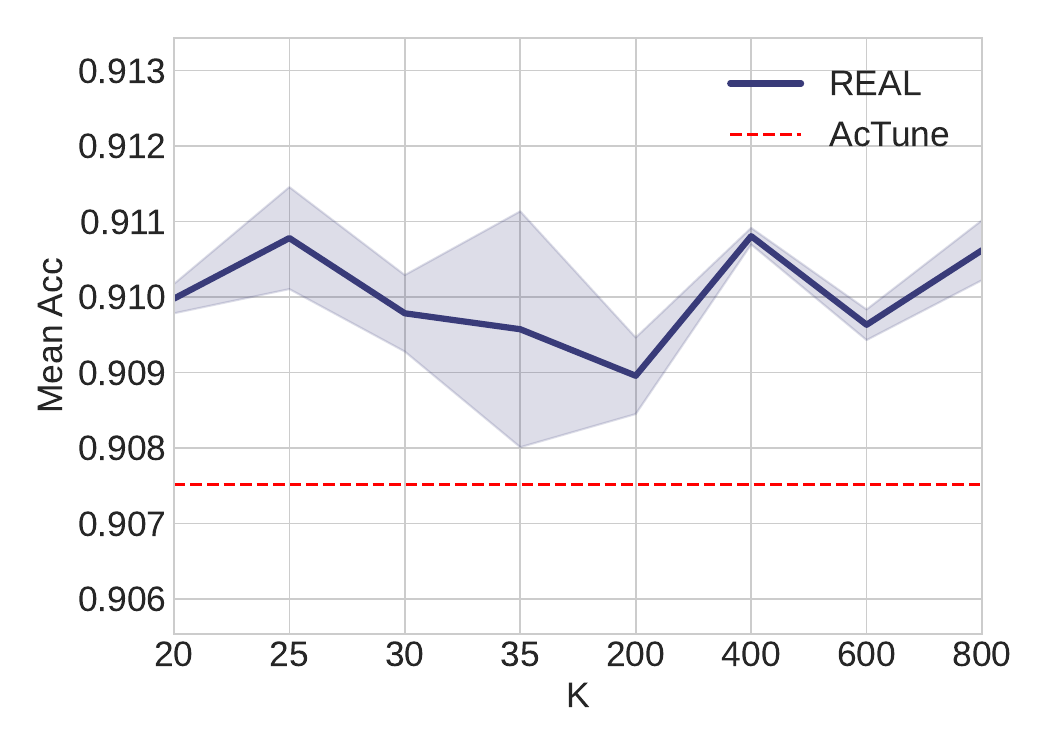}
         \caption{\ag{}}
         \label{fig:hyp_agnews}
     \end{subfigure}
     \hfill
     \begin{subfigure}[t]{0.45\textwidth}
         \centering
         \includegraphics[width=\textwidth]{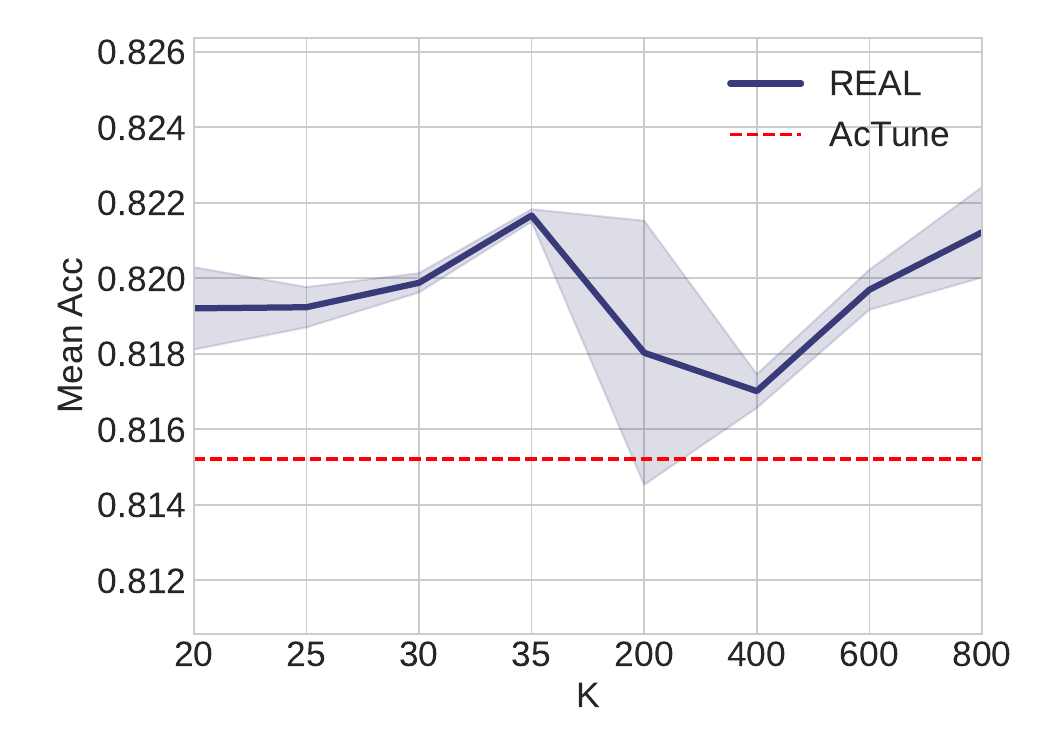}
         \caption{\pub{}}
         \label{fig:hyp_pubmed}
     \end{subfigure}
\caption{Mean acc under a wide range of \#clusters $K$ for \real{} against the best baseline \actune{} on our largest two datasets.} \label{fig:hyperparam}
\end{figure}

In this section,
we address \textbf{RQ3} by extensive ablation (Fig.~\ref{fig:ablation}) and hyperparameter studies ( Fig.~\ref{fig:hyperparam}) to understand the important components in \real{}.
\\
\textbf{A. Ablation Study.}
(1) We test \real{} with different budget allocation strategies (Eq.~\ref{eq:budget_clu}) per cluster.
 (1.1)   Ignore the idea of ``allocation by cluster''. 
Specifically, we rank all the instances in $\Du$  based on its erroneous probability (Eq.~\ref{eq:ep_ins}), and select top-$b$ instances per round (\texttt{REAL pool}).
 (1.2) For each cluster, uniformly sample $B/K$ pseudo errors, i.e., ignore cluster error weights in Eq.~\ref{eq:budget_clu}. (\texttt{REAL uniform})\\
(2) \real{} randomly  samples within each cluster's pseudo errors (line~\ref{lst:line:minekbk} in Algorithm~\ref{algo:real}) based on the adaptive budget.
Given the adaptive budget in each cluster, we also try to sample:
    (2.1) Instances with the largest erroneous probabilities in Eq.~\ref{eq:ep_ins} (\texttt{REAL cluster});
    (2.2) Pseudo errors with the largest prediction entropy (\texttt{REAL entropy}).\\
Fig.~\ref{fig:ablation} show that most of \real{}'s
variants still perform better than the best baseline \actune{} on large datasets \ag{} and \pub{}.
However, the results on \sst{} are unstable. The reason may be that the decision boundary of binary classification is too simple so that dedicated methods are not necessary.
\texttt{REAL cluster} fails only on \sst{}.
\texttt{REAL uniform} is slightly worse than \actune{} in later rounds, which indicates the importance of weighted budget allocation for \real{}. 
\texttt{REAL pool}  is very close to \actune{} on \pub{}, possibly because lacking diversity hurts it on the most difficult dataset. \\
\textbf{B. Hyperparameter Study}
We study the impact of varying the number of clusters $K$.
Experiment results in  Fig.~\ref{fig:hyperparam} shows our method stably beats the best baseline  across a wide range of $K$ (on a scale of tens to hundreds).

\section{Conclusion and Future Work}
We present \real{}, a novel \al{} sampling  strategy that selects \emph{representative pseudo errors} for efficient model training.
We define pseudo errors as minority predictions within each cluster.
The sampling budget per cluster is adaptive to the cluster's total estimated error density. Experiments on five datasets demonstrate that  \real{} performs better than other AL sampling  strategies consistently. By analyzing the actively sampled instances, we find that \real{} improves over all the best-performing baselines by guiding uncertainty sampling in errors near the decision boundary.
The ablation study shows most alternative designs of \real{} still beat the state-of-the-art baseline.

Future work will investigate the theoretical effectiveness of selecting errors near decision boundary for \al{} and the diversity within pseudo errors.
Currently we only take  text classification as an example to illustrate the effectiveness of \real{}.
But the framework of \real{} can be easily adapted to other tasks such as image classification implemented in neural classification architectures.

\section*{Ethical Statement}
All the datasets are widely-used benchmark text classification datasets and are publicly-available online, which do not have any privacy issues.
Also, our approach can benefit data labeling workers and bring welfare to them.
Data labeling is very costly and labour-intensive. 
For example, labeling toxic content is reported to be a ``mental torture''~\cite{perrigo2022inside}.
Our approach aims to make active learning more label-efficient and can reduce the workload of data labeling workers, which is beneficial to the mental health of data labeling workers.

\section*{Acknowledgments}
This work was done during Cheng Chen's internship at Singapore Management University (SMU) under the supervision of of Dr. Yong Wang.
This work was supported by
the National Key Research and Development Program of China (2020YFB1710004),
Lee Kong Chian Fellowship awarded to Dr. Yong Wang by SMU,
and the National Science Foundation of China under the grant 62272466.
We would like to thank all the anonymous reviewers for their valuable feedback.

%
%
%
\bibliographystyle{splncs04}
\bibliography{AL}

\end{document}